\documentclass[conference,a4paper]{IEEEtran}
\usepackage{amssymb}
\usepackage{verbatim}
\usepackage{pdfpages}
\usepackage{graphicx}
\usepackage{changepage}
\usepackage{enumerate}
\usepackage[numbers,sort]{natbib}

\usepackage{color}

\ifCLASSINFOpdf
\else
\fi
\usepackage{bm}
\usepackage[cmex10]{amsmath}
\usepackage{array}

\usepackage[pagebackref=true,breaklinks=true,colorlinks,bookmarks=true]{hyperref}

\usepackage[switch]{lineno}
\begin{document}
%
\title{\LARGE Boundary-based Image Forgery Detection by Fast Shallow CNN}

\author{\IEEEauthorblockN{
Zhongping Zhang, Yixuan Zhang, Zheng Zhou, Jiebo Luo}
\IEEEauthorblockA{Department of Computer Science, University of Rochester, Rochester, NY 14627\\
\ \{zzhang76, yzh215, zzhou31\}@ur.rochester.edu, jluo@cs.rochester.edu}}


%


\maketitle

\begin{abstract}
Image forgery detection is the task of detecting and localizing forged parts in tampered images. Previous works mostly focus on high resolution images using traces of resampling features, demosaicing features or sharpness of edges.
However, a good detection method should also be applicable to low resolution images because compressed or resized images are common these days. To this end, we propose a Shallow Convolutional Neural Network (\textsl{SCNN}) capable of distinguishing the boundaries of forged regions  from original edges in low resolution images. SCNN is designed to utilize the information of chroma and saturation. Based on SCNN, two approaches that are named Sliding Windows Detection (\textsl{SWD}) and Fast SCNN respectively,  are developed to detect and localize image forgery region.
Our model is evaluated on the CASIA 2.0 dataset. The results show that Fast SCNN performs well on low resolution images and achieves significant improvements over the state-of-the-art.

\end{abstract}


%
\IEEEpeerreviewmaketitle

\section{Introduction}

\begin{figure}[!t]
\centering
\includegraphics[width=3.2in]{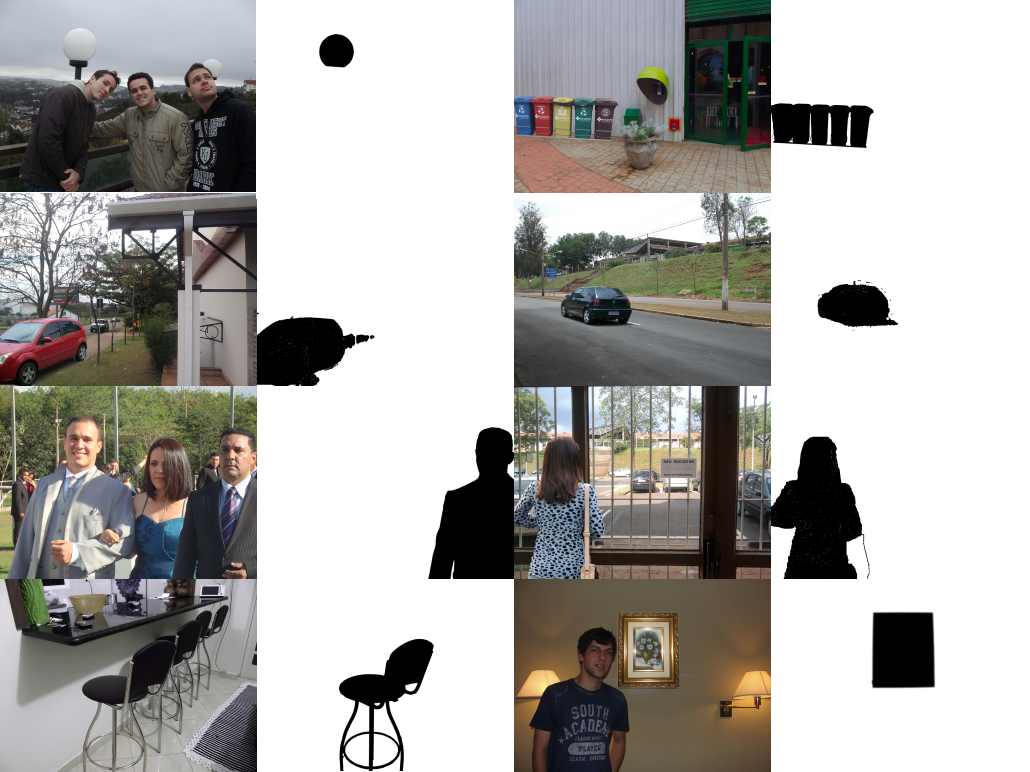}
\caption{Overview of image forgery detection. The first and third columns show example tampered images. The second and fourth columns show the corresponding ground truth.}
\label{fig_overview}
\vspace{-0.5cm}
\end{figure}

With the popularity of cameras and image processing apps, forged images become quite common in social media. Therefore, it is important to develop an effective method to detect and localize the forged region in an image, especially when images are used as critical evidence. The task of detecting and localizing forged region in an image is referred to as image forgery detection. As shown in Fig. \ref{fig_overview}, given a tampered image, the goal of image forgery detection is to predict the ground truth image of the forged region.

Over the past decade, researchers have focused on detecting the traces of resampling features (e.g. \cite{gallagher2008image}, \cite{mahdian2008blind}, \cite{popescu2005exposing}). Such methods exploits the specific statistical changes during the process of interpolation, but can only work on original TIFF or JPEG images, because compression or resizing can further change the resampling features. In recent years, with the successful application and constant perfection of deep learning, some approaches (e.g. \cite{bappy2017exploiting}, \cite{bunk2017detection}) are proposed based on Convolutional Neural Networks. However, many of these approaches can only be applied to high resolution images because they rely on resampling features or sharpness of edges. 

In order to propose an approach that can generalize well on low resolution images, we compare and contrast many pairs of low resolution images. We observe the difference between the tampered part and original part primarily in their chroma and saturation. Based on this observation, which is also consistent with \cite{wang2009effective}, we introduce SCNN to classify and distinguish forgery boundaries and original edges. Our SCNN contains two convolutional layers and two fully-connected layers. Furthermore, we discard pooling layers, and limit the number of filters and epochs. The main reason we apply a shallow network here is that we want the network to only focus on information of edges. As described in \cite{zeiler2014visualizing}, CNN often extracts information from edges in the earliest layers. If more layers are used, the network tends to learn extra information, such as shape and position. Several other key reasons are further discussed in Section~\ref{methods_SCNN}.

\begin{figure*}[!t]
\centering
\includegraphics[width=7.2in]{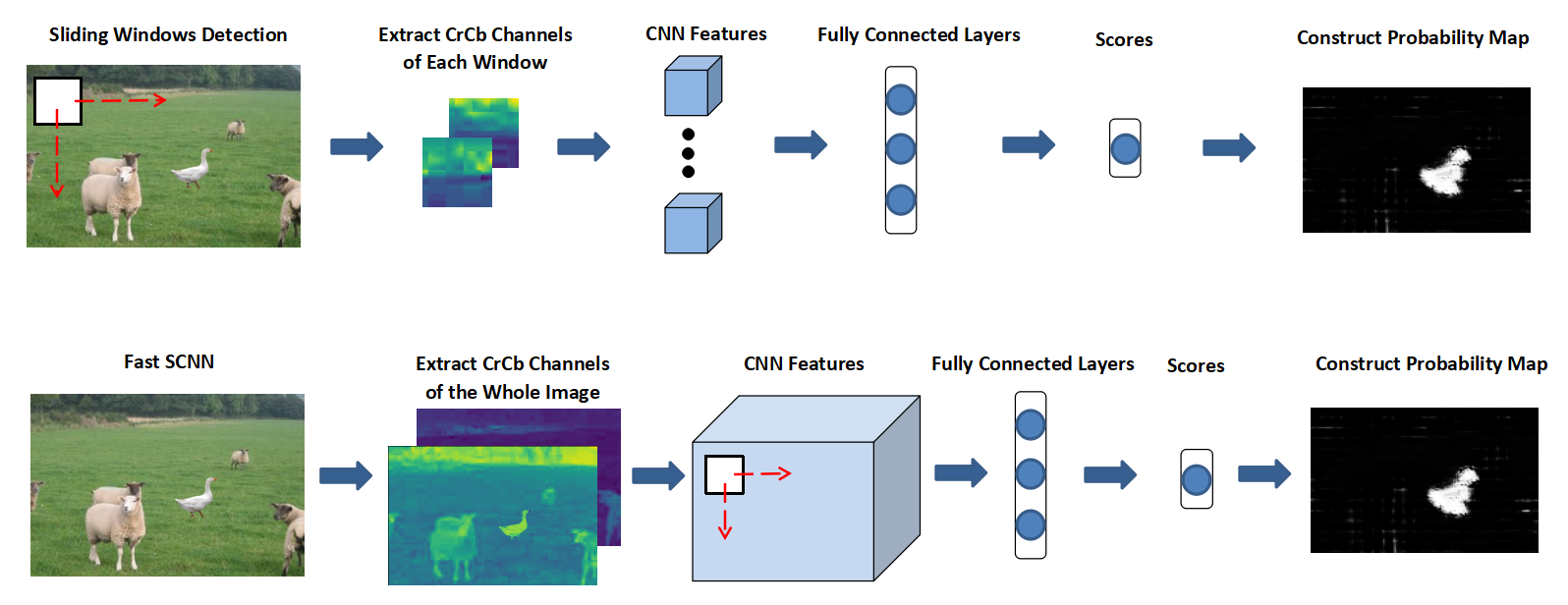}
\caption{The frameworks of SWD and Fast SCNN. SWD extracts patches from the original image. Fast SCNN extracts patches from CNN features to avoid redundant computation.
}
\label{fig_structure}
\vspace{-0.3cm}
\end{figure*}

While the proposed SCNN can be used to detect tampered edges, another challenging task is localization. To make SCNN work on an entire image, our first approach, SWD, is to apply a 32*32 window sliding over the whole image. At each step of the window, a confidence score is calculated and the probability map of the whole image is constructed by these scores. After thresholding and median filtering on the probability map, the final result for localizing the forged part of the image is generated. However, one downside of this method is time-consuming. It is infeasible to apply SWD to images of large sizes due to its high  computation cost. Inspired by the Fast RCNN \citep{girshick2015fast}, we propose an approach named Fast SCNN that is faster and more efficient than SWD. The proposed Fast SCNN computes CNN features of the whole image. The features are then passed to the fully connected layers of SCNN. Fig. \ref{fig_structure} shows the frameworks of both SWD and Fast SCNN.

Our experiments are mainly conducted on the CASIA 2.0 \citep{dong2011casia} dataset. Measurements include patch-level classification, image forgery detection and result analysis. We show that SCNN performs well on patch classification and Fast SCNN outperforms the state of the art. Through result analysis, we conclude that SCNN captures the contrast of chroma and saturation around edges rather than the shape of objects.

Our main contributions can be summarized as follows:
\begin{enumerate}[1.]
	\item We introduce an effective method for generating edge patches from images. The method of patch extraction strongly influences the performance of SCNN.
	\item We propose SCNN, a network which is shallow yet effective for classifying forgery boundaries and original edges.
	\item Based on SCNN, we further propose Fast SCNN, an efficient approach that can be applied to low resolution image forgery detection.
\end{enumerate}

\section{Related Work}
Image forgery detection has a long history and many algorithms have been developed. These algorithms can be divided into two main categories: signal processing based methods and deep learning based methods. There are also some novel methods that include metrology detection or demosaicing traces detection. We will briefly discuss them along with signal processing based methods.

\textbf{Signal processing based methods.} Earlier image forgery detection research is focused on signal processing. Therefore, many methods are proposed using interpolation marker, e.g. \cite{gallagher2005detection}, \cite{popescu2005exposing}, \cite{popescu2005exposing2}, \cite{prasad2006resampling}, \cite{mahdian2008blind}. In \cite{gallagher2005detection}, Gallagher et al. exploit the second order derivative of image to expose the periodicity in the variance function generated by interpolation. A major downside of this method is that it cannot be applied to rotated or skewed images. Based on this, an improved approach is developed by Mahdian et al. \cite{mahdian2008blind}. They introduce the Radon transform and auto-covariance to make the algorithm suitable for rotated or skewed images. Although the method is more generalized, it still focuses on uncompressed images. Luo et al. \cite{luo2010detecting} propose a framework for detecting the tampered region in compressed images. Their method first decompresses images and then evaluates the interpolation marker. However, with the development of compression techniques in recent years, it can be difficult to detect the compression format of an image and decompress the image correspondingly without additional information. In \cite{gallagher2008image}, since images from digital cameras contain traces of resampling as a result of demosaicing algorithms, such traces of demosaicing is employed to detect the forged parts of images. Although the approach is reliable for TIFF or JPEG compressed images, it may fail on other formats of compressed images. Wu et al. \cite{wu2012detecting} use a metrology method to infer the forged regions. Although it can be applied to different image formats, one major limitation is that the method is dependent on a particular geometric relationship to recover the camera calibration. In summary, signal processing based methods are mostly applied under specific circumstances.

\textbf{Deep learning based methods.} In recent years, researchers begin to exploit deep learning based models, e.g. \citep{rao2016deep}, \citep{zhang2016image}, \citep{bayar2016deep}, \citep{bunk2017detection}, \citep{bappy2017exploiting}, \citep{bayar2017robustness}.  Many deep learning based methods \citep{bappy2017exploiting}, \citep{bayar2016deep}, \citep{bayar2017robustness} still utilize the resampling features in image forgery detection. Therefore, the major weaknesses of these methods are similar to signal processing based methods. In an effort to be independent from resampling features, methods in \citep{rao2016deep} and \citep{zhang2016image} directly apply the neural network on the original images. Rao et al. \citep{rao2016deep} attempt to use a 10-layer CNN and Zhang et al. \citep{zhang2016image} use an autoencoder to perform binary classification. However, their methods are prone to overfitting on patch datasets and we will explain the reason in Section~\ref{sec:exp}. In \citep{bappy2017exploiting}, Bappy et al. employ a hybrid CNN-LSTM-CNN model to capture discriminative features between the tampered and original regions. Their method focuses on the difference between edges, especially the difference of sharpness. While the sharpness of edges is a good indicator to classify tampered regions in high resolution images, it is not effective in low resolution images that are rather smooth.

In this paper, we propose SCNN as the main component for image forgery detection. We force SCNN to learn the changes in chroma and saturation by combining the network with signal processing methods. To localize forged regions, we develop SWD and Fast SCNN that both perform well on low resolution images while Fast SCNN is more computationally efficient.

\section{Methods}
\label{methods}
In this section, we describe the method of extracting patches, and the frameworks of SCNN and Fast SCNN. Additionally, we discuss the implementation details and the reasons for Fast SCNN.

\subsection{Patch Extraction}
To construct a patch dataset, we first generate the ground truth by subtracting the tampered images from their corresponding original images. The ground truth can help us pick out tampered regions. Patches are extracted by sliding windows on tampered images. To suppress the influence of shape and position, the window size is limited to 32*32 and the stride is equal to 10. Patches where the tampered region occupies more than 35 percent and less than 65 percent of the patch area are categorized as tampered {\it boundary patches} (note that we are only interested in detecting forgery region boundaries). Patches without any tampered region are categorized as normal patches. From the CASIA 2.0 dataset, we extract 60671 normal patches and 43166 tampered patches from 4000 pairs of images. The remaining images are used as the test data for conducting image forgery detection experiments.
\label{patch_extraction}


\subsection{SCNN}
\label{methods_SCNN}
In our work, we want to lead SCNN to discriminate the changes in chroma and saturation. Inspired by \citep{wang2009effective}, the input of SCNN is first converted from the RGB space to YCrCb color space. To exclude the illumination information, we pass only CrCb channels in the convolution layers. The CrCb channels are calculated by   
\begin{equation}
Y = K_R\cdot R+K_G\cdot G + K_B\cdot B
\end{equation}
\begin{equation}
C_R = \frac{1}{2}\cdot\frac{R-Y}{1-K_R}
\end{equation}
\begin{equation}
C_B = \frac{1}{2}\cdot\frac{B-Y}{1-K_B}
\end{equation}
where $K_R$, $K_G$ and $K_B$ are constants corresponding to R,G and B channels, respectively, and required to satisfy the constraint $K_R+K_G+K_B=1$. 

The CrCb information is then fed to the convolutional (\textit{Conv}) layers of SCNN. Here we employ only two convolution layers for the following reasons. First, SCNN is designed to classify tampered edges. According to \citep{zeiler2014visualizing}, CNN extracts edge information in the earliest layers. In order to prevent SCNN from learning complex spatial features from edges, we limit the number of Conv layers to 2. Another reason is that SCNN needs to be applied many times to localize the tampered region in a single image. Therefore, a deep neural network would take an unacceptable amount of time.


In addition, we randomly initialize 20\% of the filters (i.e., convolution kernels) with a Laplacian kernel:
\begin{equation}
h(x,y)=
 \left[
 \begin{matrix}
   0 & 1 & 0 \\
   1 & -4 & 1 \\
   0 & 1 & 0
  \end{matrix}
  \right] \tag{4}
\end{equation}
The Laplacian kernel can expose both edge and resampling features of an image. It is used to force SCNN to learn information from edges and resampling features. 

Another modification in SCNN is discarding pooling layers. Since SCNN is quite shallow, we discard the pooling layers to preserve as much information as possible. The exact network configuration of SCNN is shown in Table \ref{SCNN_structure}.

In summary, SCNN is based on the following four modifications on a typical CNN:
\begin{enumerate}[1)]
	\item Extracting the CrCb channels from RGB space
    \item Limiting the number of Conv layers
    \item Initializing some of filters with a Laplacian kernel
	\item Discarding the pooling layers  
\end{enumerate}

\begin{table}
\caption{ SCNN architecture for CASIA 2.0. }
\label{SCNN_structure}
\centering
\begin{tabular}{c|c|c}
\hline
Layer(type) & Shape & Param\\
\hline
Input & (None, 32, 32, 3) & - \\
\hline
CrCb Channels & (None, 32, 32, 2) & - \\
\hline
Conv1 & (None, 30, 30, 32) & 608\\
Activation1 ('ReLU') & (None, 30, 30, 32) & 0\\
\hline
Conv2 & (None, 28, 28, 32) & 9248\\
Activation2 ('ReLU') & (None, 28, 28, 32) & 0\\
\hline
Flatten & (None, 25088) & 0\\
\hline
Dense1 & (None, 64) & 1605696 \\
Activation3 ('ReLU') & (None, 64) & 0\\
\hline
Dropout & (None, 64) & 0\\
\hline
Dense2 & (None, 1) & 65\\
Activation4 ('Sigmoid') & (None, 1) & 0\\
\hline
\end{tabular}
\end{table}

\subsection{Fast SCNN}

\textbf{Sliding Window Detection.} Assuming SCNN is trained, a feasible method of localizing the tampered image regions is Sliding Window Detection (\textsl{SWD}). We start by picking a certain window of an image, feed the window into SCNN and compute a confidence score to predict whether it is tampered. The confidence score is stored in the corresponding position of a probability map. Then the window slides over and outputs another confidence score. After sliding the window through the entire image, a complete probability map is constructed. The main weakness of this method is slowness, because of redundant computations. Therefore, we propose Fast SCNN to alleviate the problem.

\textbf{Fast SCNN.} 
Compared with SCNN that takes as input a 32*32 patch, Fast SCNN takes the entire image as the input. It first produces feature maps by processing the entire image with Conv layers. Supposing the input shape of the image is $(n,m,3)$, the dimension of the feature maps is $(n-4,m-4,32)$. We extract feature vectors with dimension $(28,28,32)$ from feature maps and feed them into fully-connected layers. The parameters of Fast SCNN are all trained by SCNN on the patch dataset. At this time, redundant computation of the Conv layers is avoided.
\label{methods_Fast}

\section{Experiments}
\label{sec:exp}
In this section, we first present experiments on patch classification and image forgery detection on the CASIA 2.0 dataset.  The CASIA 2.0 dataset contains 7491 authentic and 5123 tampered images in various kinds of format, such as JPEG, BMP and TIFF. Compared with other datasets \citep{IEEEdataset} and \citep{nimble2016datasets}, it is more challenging because it contains mostly low resolution images and introduces post-processing on the boundary area of tampered regions. We then analyze the results to find out some specific information that SCNN learns.

\subsection{Patch Classification}
\begin{figure}
\centering
\noindent\makebox {
\begin{tabular}{cccc}

\includegraphics[width=0.18\linewidth]{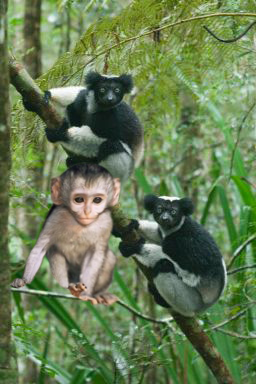} &
\includegraphics[width=0.18\linewidth]{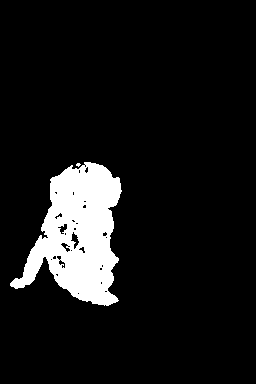} &
\includegraphics[width=0.18\linewidth]{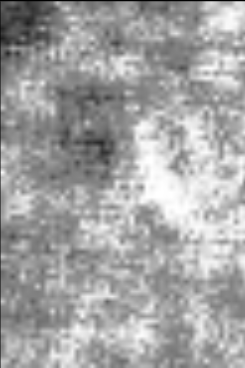} &
\includegraphics[width=0.18\linewidth]{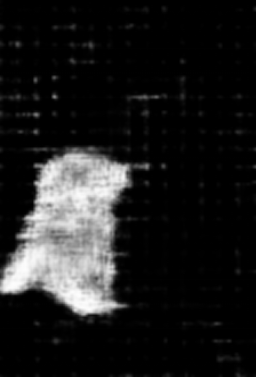} \\
 a) & b) & c) & d) \\

\end{tabular}
}
\caption{Comparison of SCNN and CNN-10. (a) Original image (b) Ground truth (c) Probability map constructed by CNN-10 d) Probability map constructed by SCNN.}
\label{fig:CNN-10}
\vspace{-0.3cm}
\end{figure}

\begin{figure}[!t]
\centering
\includegraphics[width=3.2in]{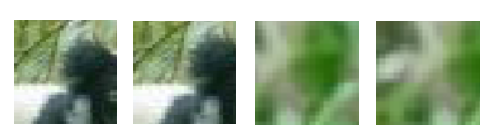}
\caption{Two pairs of patches. They are extracted from the original image in Fig \ref{fig:CNN-10}, but their positions are slightly different. Outputs of CNN-10 are 0.991, 0.001, 0.896 and 0.000, respectively.
}
\label{fig_compare}
\vspace{-0.2cm}
\end{figure}

The patch dataset which is used to train SCNN is constructed as the method described in Section~\ref{patch_extraction}. We apply various classification models on the patch dataset. Accuracy results are shown in Table \ref{accuracy}. From Table \ref{accuracy}, we can see that complex models, Inception-v4 and ResNet-50, suffer seriously from overfitting. The reason is that ResNet-50 and Inception-v4 are too complex to classify edges effectively. CNN-10\footnote{Since we use different methods to extract patches, the accuracy of CNN-10 in Table \ref{accuracy} is different from the original paper.}  performs better than SCNN on patch classification. However, when we apply CNN-10 on entire images, it generates almost all-white probability maps as Fig. \ref{fig:CNN-10} shows. To find out the reason, we select some patches from the same pictures and compare their outputs. As we can see in Fig. \ref{fig_compare}, slightly changing positions will cause totally different output. The reason is that CNN-10 is still too deep for edge information. It learns useless information, such as position, shapes and luminance. This information causes CNN-10 to be unstable and unreliable on entire images. Therefore, we apply SCNN as the main classification component for the image forgery detection task.

\begin{table}[!t]
\caption{Results of patch classification for CASIA 2.0.}
\label{accuracy}
\centering
\begin{tabular}{|c|c|c|c|c|}
\hline
Model (type) & SCNN & CNN-10 \citep{rao2016deep} & Inception-v4 & ResNet-50\\ 
\hline
Accuracy & 80.41\% & 90.72\% & 61.86\% & 56.70\% \\
\hline
\end{tabular}
\end{table}

\begin{figure}[!t]
\centering
\includegraphics[width=2.82in]{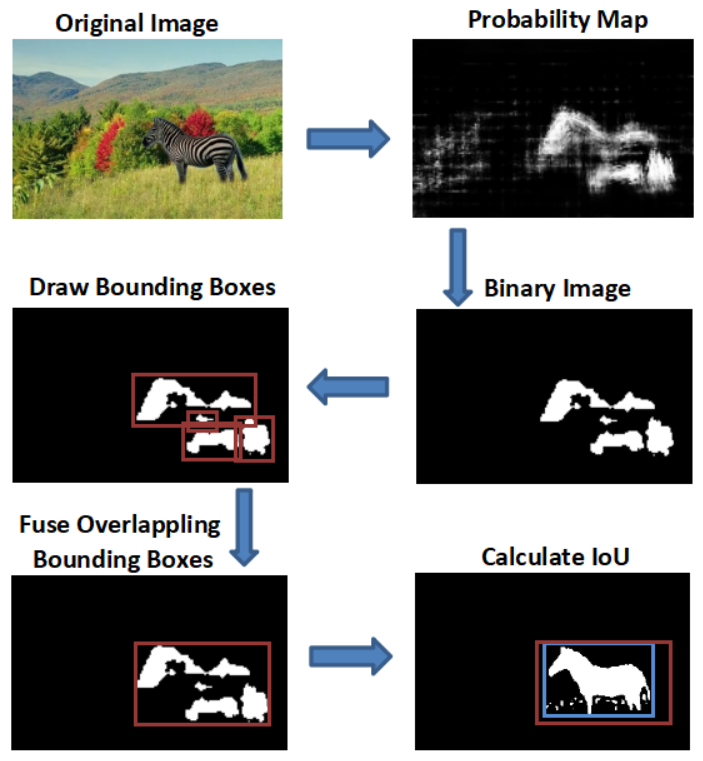}
\caption{ The process of generating a tampered region bounding box.
}
\label{fig:IoU}
\vspace{-0.2cm}
\end{figure}

\subsection{Localizing Forged Image Regions}
One of the main goals of image forgery detection is to localize tampered image regions. Recall that we have two methods to localize a tampered region. We first compare the performance of the two models in terms of time. Then we compare Fast SCNN with the-state-of-the-art models that  can be applied to low-resolution images.

\textbf{Computation time.} We evaluate the computation time of SWD and Fast SCNN on images of different sizes. The results are shown in Table \ref{table:time}. In our experiment\footnote{All timing information is on a Geforce GTX 1070 overclocked to 2000 MHz.}, the stride is set to 2 and both models are tested on the same images. As Table \ref{table:time} shows, Fast SCNN takes only about 60\% time of SWD. Furthermore, the time difference between Fast SCNN and SWD increases as the image size increases. 

\begin{table*}
\caption{Accuracy results of each category in CASIA 2.0.}
\label{table:acc}
\centering
\begin{tabular}{|c|c|c|c|c|c|c|c|c|c|c|}
\hline
Category & Animal & Architecture & Article & Character & Indoor & Nature & Plant & Scene & Texture & Average\\ 
\hline
Accuracy Based on IoU (\%) & 89.19 & 91.30 & 85.28 & 76.47 & 84.21 & 81.25 & 67.35 & 89.90 & 63.23 & 80.91 \\
\hline
Accuracy Based on Human Judgment (\%) & 90.32 & 94.85 & 91.92 & 77.36 & 90.72 & 89.69 & 68.04 & 90.43 & 63.92 & 84.14\\
\hline
\end{tabular}
\end{table*}

\textbf{Accuracy.} 
Although the accuracy can be measured by whether the probability map can help localize the tampered part correctly, we want to develop a more concrete method that does not depend on human subjectivity. Therefore, a measurement method based on IoU is proposed. We first process the probability map by binary thresholding and median filtering. Then we draw bounding boxes around bright parts and combine overlapping bounding boxes into one. We compare the generated bounding box with the ground truth bounding box. If IoU$>$0.5, we consider the result as correct. Fig. \ref{fig:IoU} shows the process of generating a bounding box. The localization accuracy of each category is reported in Table \ref{table:acc}. From Table \ref{table:acc}, we see that our model can effectively localize image forgery regions. In Table \ref{table:acc}, the accuracy based on IoU is lower than the accuracy based on human judgment especially in the  "Nature" category. It is because IoU cannot represent the tampered region well on images in which the forged part is the background, such as sky, river, etc. Table \ref{table:acc} demonstrates that the model performs well on most categories except "character", "plant" and "texture". We analysis the reasons in Section \ref{experiments: res}.

\begin{figure}
\centering
\noindent\makebox {
\begin{tabular}{cccc}

\includegraphics[width=0.18\linewidth]{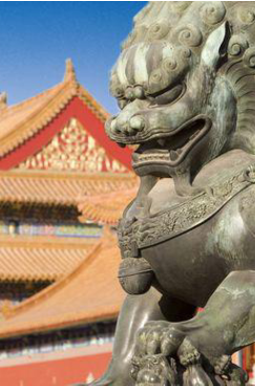} &
\includegraphics[width=0.18\linewidth]{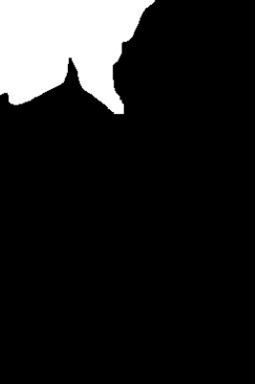} &
\includegraphics[width=0.18\linewidth]{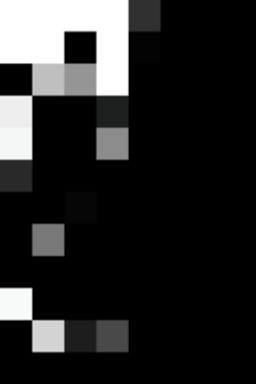} &
\includegraphics[width=0.18\linewidth]{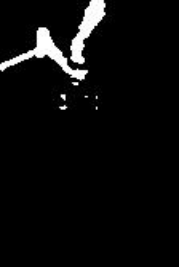} \\
 a) & b) & c) & d)\\
\end{tabular}
}
\caption{Comparison of Fast SCNN and Zhang et al. \citep{zhang2016image}. (a) Original Image, (b) Ground truth (the bright part is forged), (c) Output of Zhang et al. \citep{zhang2016image}, (d) Output of Fast SCNN (which marks the forgery boundaries). }.
\label{fig:Compare Singapore}
\vspace{-0.3cm}
\end{figure}

Comparing Fast SCNN with the-state-of-art models, we report the accuracy results\footnote{Accuracy of the first three models on CASIA 2.0 are obtained from \citep{zhang2016image}} in Table \ref{table:accuracy}. Table \ref{table:accuracy} shows that our method is more general in terms of image format than the methods of Bianchi et al. \citep{bianchi2012image} and Thing et al. \citep{thing2012improved}. Since Fast SCNN does not utilize interpolation features, it can be applied to all image formats instead of being constrained to JPEG images. Furthermore, our model is a pixel-based model rather than a patch-based model. Fig. \ref{fig:Compare Singapore} shows the different outputs between Fast SCNN and the model of Zhang et al. The algorithm of Zhang et al. tends to consider the entire sky region as the tampered region. This is unreasonable because the sky is tampered in this image but is real in its original image (note that their method performs classification on a patch by patch basis). In other words, their method blindly learns this kind of sky as tampered as their method classifies tampered regions using information of particular color or shape. Furthermore, their result, 87.51\%, is the accuracy of patch-level classification. It is not designed to accurately localize tampered regions and  the IoU-based accuracy is expected to be poor.  

\begin{figure*}[!t]
\centering
\includegraphics[width=6.0in]{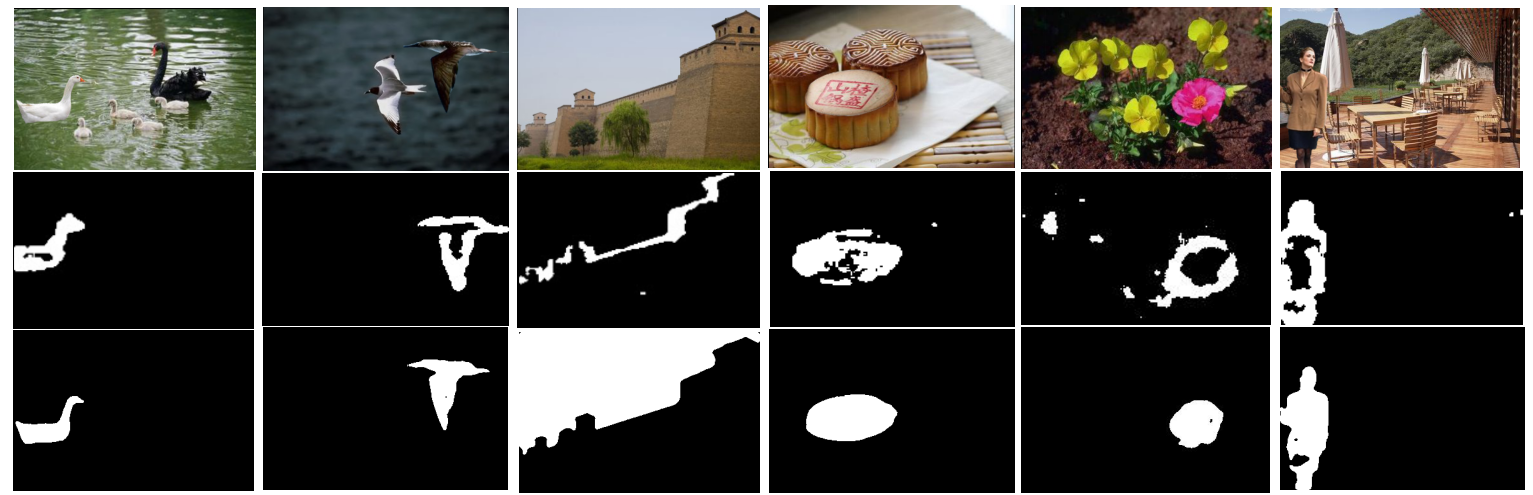}
\caption{Image examples. Top row: original images; Second row: binary probability maps by Fast SCNN; Third row: the ground truth.
}
\label{fig:good}
\vspace{-0.2cm}
\end{figure*}

To make our discussion more intuitive, we include several examples in Fig. \ref{fig:good} to show the efficacy of Fast SCNN.

\begin{table}
\caption{Computation time of SWD and Fast SCNN.}
\label{table:time}
\centering
\begin{tabular}{|c|c|c|c|}
\hline
Image Size & SWD (s/im) & Fast SCNN (s/im) & Time Saved (s/im)\\ 
\hline
200$\sim$400 & 17.16 & 10.31 & 6.85\\
\hline
400$\sim$600 & 52.86 & 31.21 & 21.65\\
\hline
600$\sim$800 & 109.04 & 63.47 & 45.57\\
\hline
800$\sim$1000 & 234.53 & 126.27 & 108.26 \\
\hline
\end{tabular}
\end{table}


\begin{table}[!t]
\caption{Detection accuracy comparison with other models applicable to low resolution images. * indicates  patch-level accuracy.}
\label{table:accuracy}
\centering
\begin{tabular}{|c|c|c|}
\hline
Model & Accuracy (JPEG, \%) & Accuracy(TIFF, \%) \\ 
\hline
Bianchi et al. \citep{bianchi2012image} & 40.84 & -  \\
\hline
Thing et al. \citep{thing2012improved} & 79.72 & - \\
\hline
Zhang et al. \citep{zhang2016image} & 87.51* & 81.91* \\
\hline
Fast SCNN & 85.35 & 82.93\\
\hline
\end{tabular}
\end{table}

\subsection{Result Analysis}
\label{experiments: res}
In this section, we analyze the results by comparing images to verify the functionality of our model. We hope to figure out which information is used by SCNN to differentiate tampered boundaries from original edges.

\begin{figure}[!t]
\centering
\includegraphics[width=3.0in]{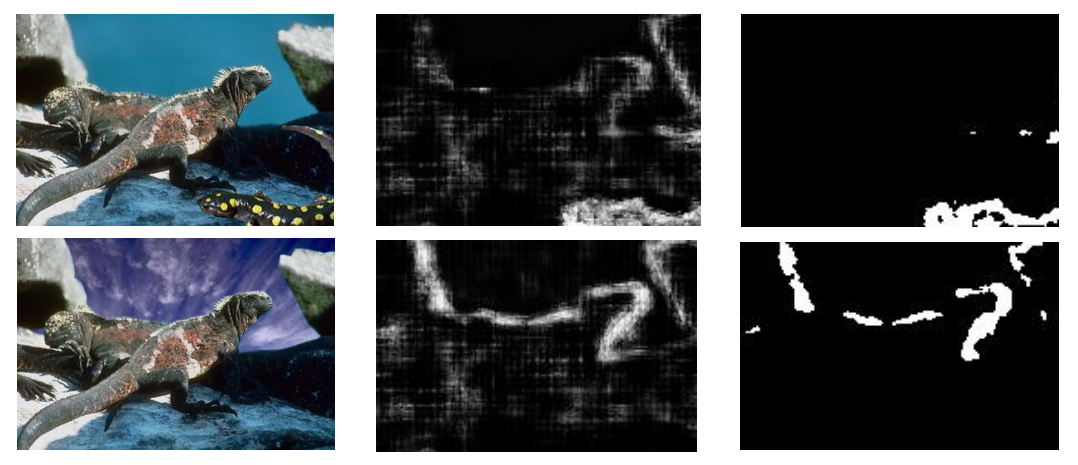}
\caption{A pair of images with different parts tampered. 
}
\label{fig:Compare}
\vspace{-0.5cm}
\end{figure}

First, we try to confirm that SCNN learns the difference between original edges and tampered boundaries rather than the shapes of particular edges. Therefore, we conduct the experiment on similar images with different parts tampered are shown in Fig. \ref{fig:Compare}. From Fig. \ref{fig:Compare}, we can see that Fast SCNN does not output the same probability map. For the image in the first row, the probability map and the binary image mark the colorful lizard added in the bottom right corner. For the image in the second row, the probability map and the binary image mark the contour of replaced sky. Therefore, it is clear that SCNN learns the difference between tampered boundaries and original edges rather than the shape of particular edges. 

\begin{figure*}[!t]
\centering
\includegraphics[width=6.0in]{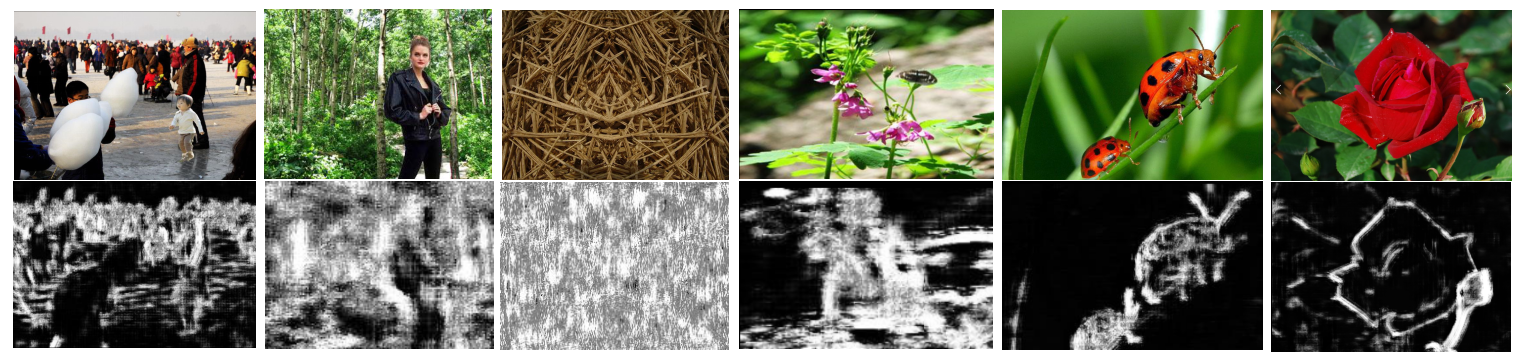}
\caption{A selection of images that were mistakenly classified and localized. Note that only the smaller laybug is forged in the 5th column. 
}
\label{fig:bad}
\vspace{-0.2cm}
\end{figure*}

Next, we try to find out why SCNN performs poorly in some categories such as "character", "texture" and "plant". Fig. \ref{fig:bad} shows a selection of images that are mistakenly classified and localized. In Fig. \ref{fig:bad}, wrong outputs mainly occur in two kinds of images. The first kind is images whose background is complex, such as the first three columns. The other kind is images that have varying focuses or depth of field effect, such as the last three columns. This indicates that SCNN not only learns the changes in chroma and saturation, but also learns the changes of focus somehow. As a result, it explains why our model performs poorly in "character", "plant" and "texture" categories. Many "character" images contain different focus, while  "texture" and "plant" images are very likely to contain  complex background.


\section{Conclusion}
In this paper, we present a novel method for image forgery detection. SCNN is used as the basic detection component to detect tampered region boundaries. Fast SCNN is proposed to detect and localize the tampered part in an image. Our model works well on all image formats, and outperforms the state-of-the-art models on low resolution images. Comprehensive experiments on the CASIA 2.0 dataset demonstrate the effectiveness of our model and confirm that our model captures abnormal changes in chroma and saturation across forged region boundaries. In the future, we will improve our model for images with complex background or different focus.




%
{
\footnotesize
\bibliographystyle{plain}
\bibliography{bibliography}
}

\end{document}